# Audio-Driven Dubbing for User Generated Contents via Style-Aware Semi-Parametric Synthesis

Linsen Song, Wayne Wu, Chaoyou Fu, Chen Change Loy, *Senior Member, IEEE,* and Ran He, *Senior Member, IEEE*

*Abstract*—Existing automated dubbing methods are usually designed for Professionally Generated Content (PGC) production, which requires massive training data and training time to learn a person-specific audio-video mapping. In this paper, we investigate an audio-driven dubbing method that is more feasible for User Generated Content (UGC) production. There are two unique challenges to design a method for UGC: 1) the appearances of speakers are diverse and arbitrary as the method needs to generalize across users; 2) the available video data of one speaker are very limited. In order to tackle the above challenges, we first introduce a new Style Translation Network to integrate the speaking style of the target and the speaking content of the source via a cross-modal AdaIN module. It enables our model to quickly adapt to a new speaker. Then, we further develop a semi-parametric video renderer, which takes full advantage of the limited training data of the unseen speaker via a video-level retrieve-warp-refine pipeline. Finally, we propose a temporal regularization for the semi-parametric renderer, generating more continuous videos. Extensive experiments show that our method generates videos that accurately preserve various speaking styles, yet with considerably lower amount of training data and training time in comparison to existing methods. Besides, our method achieves a faster testing speed than most recent methods.

## I. Introduction

WITH the popularity of the User Generated Content (UGC) [1], [2] (*e.g.* YouTube and TikTok), dubbing technologies have come into the sight of ordinary users for producing creative and entertaining contents. In this paper, we propose an automated audio-driven dubbing method for UGC production based on two requirements: 1) the dubbing method needs to handle various users; 2) most users are impatient to record a long video for training a model and wish to get dubbed videos as fast as possible.

Most existing audio dubbing methods only cope with one/several speakers, requiring massive training data and long training time (Fig. 1 (b)). Consequently, these methods are usually targeted for the Professionally Generated Content (PGC) [3], [4] production (*e.g.* films and TV shows) and rarely used in the UGC setting. Such techniques are currently inapplicable for UGC production due to the following two challenges: 1) *Speaker Variance.* Different speakers have their unique mouth shapes and textures. We define the time-varying mouth shapes of a speaker as its unique speaking style, both in terms of mouth shape and movement timing. Many methods designed for PGC production either consider only the homogeneous generation of one's video by its own audio [5] or neglect the speaking style differences between speakers [6]. Compared with PGC production, to generate high quality talking videos for any unseen speaker, the dubbing methods designed for UGC production need to preserve their unique speaking styles. The main challenge lies in tackling the speaking style of a given speaker and the speaking content of a given audio at the same time. 2) *Training Resource.* A dubbing method designed for UGC production should quickly adapt to an unseen speaker with very limited video data. For methods designed for PGC production, adapting to an unseen speaker is expensive as one needs to retrain the whole network on massive training data from the speaker. The main challenge is to generate realistic and lip synchronized talking videos rapidly after giving a short video of an unseen speaker.

In this paper, we aim at realizing audio-driven dubbing for UGC through mitigating the issue of speaker variance and lowering the consumption of training resources (Fig. 1). To tackle the aforementioned challenges, we first propose a cross-modal Adaptive Instance Normalization (AdaIN [7]) in our Style Translation Network that maps the source audio to the mouth motion of an unseen target speaker with preserved speaking style. AdaIN is popular for image style transfer [8] and we extend it to fuse information from different modalities (*i.e.* audio and video). Then, we propose a semi-parametric framework that helps to bring down the training resources, including both training data and time. The limited video

Manuscript received 26 November 2021; revised 11 June 2022 and 28 August 2022; accepted 11 September 2022. Date of publication 26 September 2022; date of current version 7 March 2023. This work was supported by the National Natural Science Foundation of China under Grant U21B2045 and Grant U20A20223. The work of Chen Change Loy was supported by the Research Innovation Enterprise (RIE) 2020 Industry Alignment Fund Industry Collaboration Projects (IAF-ICP) Funding Initiative. This article was recommended by Associate Editor Y. Wu. *(Corresponding author: Ran He.)*

Linsen Song, Chaoyou Fu, and Ran He are with the National Laboratory of Pattern Recognition, CASIA, Center for Research on Intelligent Perception and Computing, CASIA, Center for Excellence in Brain Science and Intelligence Technology, CAS, and the School of Artificial Intelligence, University of Chinese Academy of Sciences, Beijing 100190, China (e-mail: songlinsen2018@ia.ac.cn; chaoyou.fu@nlpr.ia.ac.cn; rhe@nlpr.ia.ac.cn).

Wayne Wu is with SenseTime Research, Beijing 100080, China (e-mail: wuwenyan0503@gmail.com).

Chen Change Loy is with the S-Laboratory, Nanyang Technological University, Singapore 639798 (e-mail: ccloy@ntu.edu.sg).

This article has supplementary material provided by the authors and color versions of one or more figures available at https://doi.org/10.1109/TCSVT.2022.3210002.

Digital Object Identifier 10.1109/TCSVT.2022.3210002

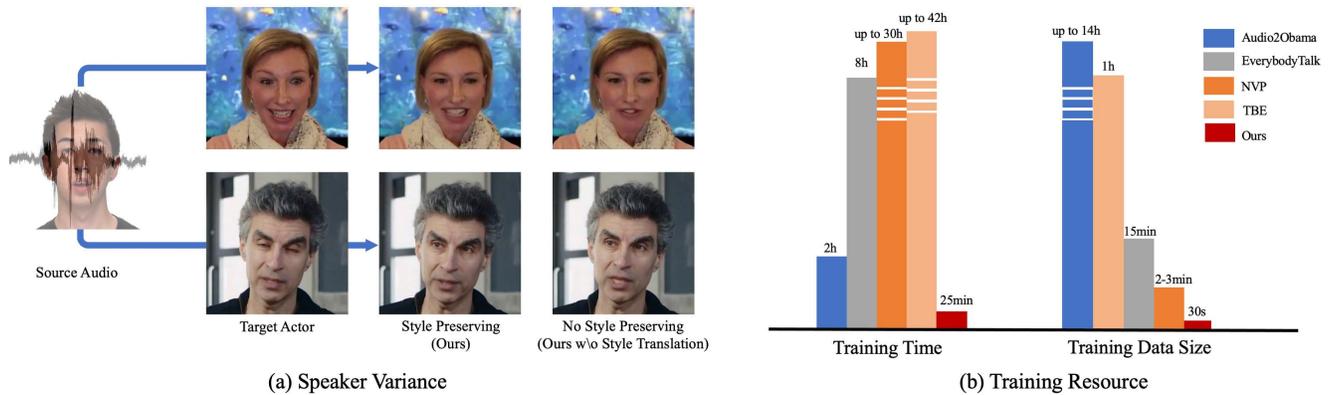

Fig. 1. **Audio-driven dubbing in User Generated Content (UGC).** Our audio-driven dubbing method tackles two challenges in User Generated Scenarios: (a) Speaker Variance and (b) Training Resource. Our method generates talking videos with speakers' unique speaking styles preserved and requires a low amount of training data and training time, which paves a way for ordinary content creators to edit talking videos in User Generated Scenarios.

data make it hard to train a generator that directly generates the mouth regions that match the specific speaker. Thus, we propose a cheaper yet effective video-level Retrieval-based Video Renderer to take full advantage of the short video data of the unseen speaker. Such a semi-parametric framework is unique since it considers video continuity. To achieve this, we extend the conventional image-level thin-plate spline (TPS) warping algorithm [9] to a video-level one with our proposed temporal regularization.

The main contributions of our work are summarized as follows:

- We present the first audio-driven dubbing framework that is designed specially to support ordinary video creators in User Generated Content (UGC) production.
- We propose a cross-modal AdaIN module in our Style Translation Network that can preserve the speaking style of a speaker while producing mouth movement that is consistent with the source audio.
- We develop a video-level Retrieval-based Video Renderer that can produce dubbed videos for an unseen speaker with very limited video data and short training time.
- We extend the image-level thin-plate spline interpolation warp algorithm with a proposed temporal regularization to improve the video continuity.

## II. RELATED WORK

### A. Audio-Driven Dubbing

Audio-driven dubbing typically refers to the technique that associates phonemes or speech features provided by the input audio with visual visemes of a target actor. Inspired by recent advances of GAN in face generation [10], [11], expression generation [12], [13] and audiovisual learning [14], [15], much progress has been made in audio-driven dubbing. Taylor et al. [16] propose to generate natural-looking speech animation with a sliding window predictor that learns a non-linear mapping from phonetic representations to active appearance model (AAM [17]) parameters. By making the learned AAM parameters independent of speakers, the method generates animation that can be retargeted to any animation rig. Karras et al. [18] design a method to learn a mapping from input waveforms to facial 3D vertex coordinates and define a trainable latent code that can be used as an emotion control parameter. Pham et al. [19] introduce a long short-term memory recurrent neural network (LSTM-RNN [20]) approach to achieve facial animation via an expression blendshape model. Different from these 3D-based schemes, methods including [21], [22], [23], [24] drive photo-realistic portraits to generate videos synchronized with input audios. However, these methods tend to produce animations of still images or cartoon-looking characters, which limits their ability in general video dubbing. With the rise of deep learning, much progress has been made in audio-driven dubbing Recently, some works focus on the generation of natural videos for audio dubbing. Suwajanakornet al. [5] train a recurrent neural network to achieve state-of-the-art synthesis results of Obama's portrait videos. This method uses up to 14 hours of Obama's speech videos to learn his mouth movements from speech audios. However, the assumptions of sharing identity between the source and the target as well as the access of long-hour training data hinder its application in many reality scenarios. Fried et al. [25] propose a transcript-based method to change the dialogue content while maintaining a seamless audio-visual flow. Despite the visual quality, it has to train separate network models for different targets and takes a relatively long time to search for mouth areas based on visemes in the provided video. Therefore, it is quite difficult to apply this method to different speakers. Yu et al. [26] design a multimodal learning method composed of two components: mouth landmark prediction and video generation. The landmark prediction component generates mouth landmarks from multimodal inputs like audio or text. The video generation component employs optical flow to model the temporal dependency of continuous frames and generates talking videos from predicted landmarks. To generate perfect video results, this method requires massive training videos for an unseen speaker (e.g., 1 hour for Donald John Trump). Song et al. [27] translate source audio into talking videos of multiple speakers by factorizing each target video frame into orthogonal parameter spaces including shape, expression and pose. Thies et al. [6] animates the 3D face model of the target speaker by speech content extracted by DeepSpeech [28], then they render facial textures and compose it with the target face image. To generate talking videos with natural head motions, recent methods focus on head pose learning from audios [29], [30] or videos [31].

### B. Visual Dubbing

Due to the distinct gap between the audio domain and the video domain, it is inherently challenging to generate plausible portrait videos directly from the input audio. Consequently, recent methods explore to first learning mouth movement representation from available videos and then complete generation or reenactment based on the learned representation. Garrido et al. [32] transfer facial motion from dubber in blend shape model, then they render the synthesized face and mouth interior into the original video. Thies et al. [33] develop an algorithm named Face2face to animate the facial expressions of the target video by a source actor. Kim et al. [34] present an approach to transfer the full 3D head position, head rotation, face expression, eye gaze, and eye blinking from a source actor to a target actor. Nagano et al. [35] design paGAN to build dynamic avatars from a single input image. Kim et al. [36] propose a visual dubbing method to maintain the distinct style of target actors when modifying facial expressions. Their video-driven method translates the facial expression of the source actor to the style domain of the target actor. The Style Translation Network in [36] is inspired by the cycle-consistent loss [37] and the trained network only supports style translation between two domains. However, these methods all perform a laborious task of generating and tracking a 3D face model, which is time-consuming and even difficult for existing footage.

Another type of approach reenacts facial portraits directly in the 2D space. Zhou et al. [38] propose an image-based visual speech animation system. In this system, a low dimensional continuous curve is used to represent a video sequence and a map from the curve to the image domain is established. A video model is trained on dense videos sampled from the talking video segments for efficient time alignment and motion smoothing of the final video synthesis. In the final video synthesis, Poisson blending [39] is applied in stitching the synthesized mouth-chin area into a background sequence. Garrido et al. [40] work on automatic face reenactment by applying a matching & transfer pipeline to transfer inner face region from source actor to target actor. Geng et al. [41] present a warp-guided generative model to activate an input photo with a set of facial landmarks. Wiles et al. [42] introduce a network named X2Face to control a source face with estimated expression parameters. Qian et al. [43] develop Additive Focal Variational Auto-encoder that can manipulate high-resolution face images in a weakly supervised manner. Although these methods generate decent frames but suffer from relatively poor temporal continuity. Tu et al. [15] design a versatile model called as FaceAnime that generates a face video from a single face image. This method first predicts 3D face dynamics from a still image by a LSTM model. Then the "imagine" face dynamics are used to generate an animated face by the proposed video generation network. As an end-to-end video-based facial reenactment network, ReenactGAN [44] transforms the source face's boundary to the target's boundary in a boundary latent space and then generates the reenacted target face with a decoder. However, the transformer and the decoder in ReenactGAN are person-specific, which limits the generalization of the model.

### C. Style Translation & AdaIN Layer

In general, style translation aims at transferring the style between samples in the same modality. In computer vision, the style can be defined as the overall color tone of images. For example, Park et al. propose a model [8] to generate landscape images with referred styles conditioned on semantic maps. Karras et al. develop a model [45] to generate a face image with a controllable style defined by the referred face image and address characteristic artifacts in the improved work [46]. In voice conversion, the style can be defined as the unique voice tone of the speaker, and voice conversion is to modify the voice tone of different speakers while keeping the speaking content unchanged. Chou et al. propose a voice conversion model [47] that respectively extracts speaker style from the target speaker and speaking content from the source speaker. Then, the voice conversion model fuses them and generates the audio of the target speaker who speaks the sentences that appeared in the audio of the source actor. The above models [8], [45], [47] apply the AdaIN layer to transfer different "style". The style information in image style translation [8], [45] and voice conversion [47] is a kind of "global information" or low order statistics. We are inspired that the AdaIN layer might work in transferring the speaking style embedded in the mouth movement.

## III. METHODOLOGY

The proposed audio-driven dubbing method is illustrated in Fig. 2. To tackle the speaker variance challenge, we design a cross-modal AdaIN module in the Style Translation Network to explicitly combine the speaking style of the target speaker and the speaking content of the source audio (Sec. III-A). To generate continuous talking videos for unseen speakers with limited video data, we develop the semi-parametric Retrieval-based Video Renderer with our proposed Temporal Warp module to improve the video continuity (Sec. III-B). Finally, we depict the scheme and details of training our network on a new speaker (Sec. III-C).

### A. Style Translation Network

The synchronization between the source audio and the generated video is the primary purpose to achieve a dubbing method. We propose a combination of three modules, *i.e.*, Content Encoder, Style Encoder and Cross-Modal AdaIN, to achieve this goal. Firstly, we design a Content Encoder to learn the speaking content embeddings that are strongly related to the speaking sentences. Then, for a given unseen speaker whose speaking style is unpredictable, we design a Style Encoder to explicitly learn its speaking style embeddings. Finally, since we need to drive an unseen speaker with the audio of another speaker in the UGC production, we propose a cross-modal AdaIN module in the Decoder to combine the extracted speaking content and style embeddings to get the inferred landmarks. The inferred landmarks preserve the speaking style of the target speaker and synchronize with the source audio.

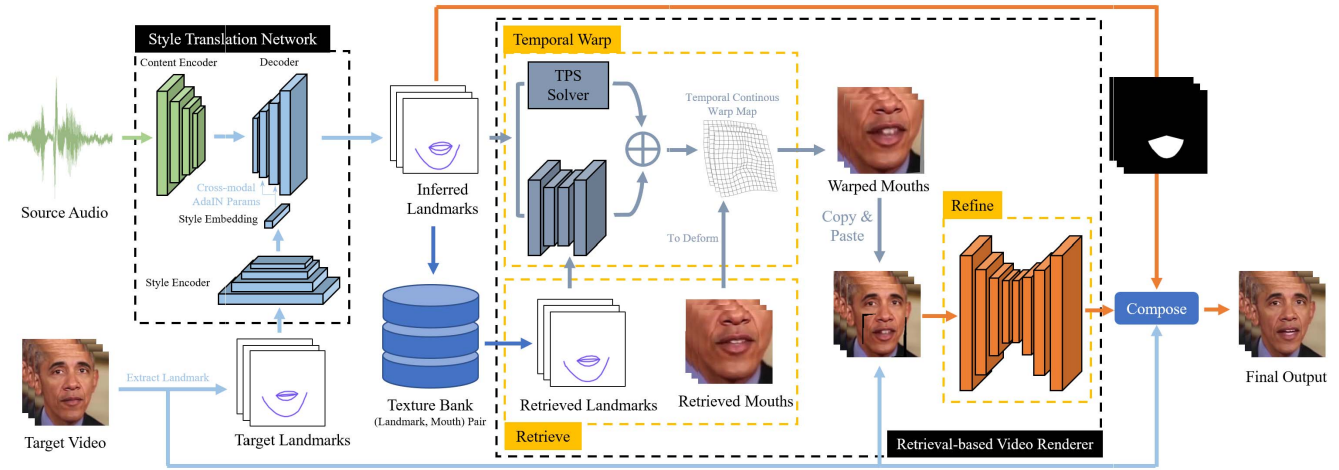

Fig. 2. **Architecture.** Our method contains an audio-visual Style Translation Network and a Retrieval-based Video Renderer. Integrating with the proposed cross-modal AdaIN layer, the Style Translation Network infers mouth movement from the source audio and preserves the speaking style of the target actor. The Retrieval-based Video Renderer learns to generate photo-realistic videos from inferred landmarks through a series of modules including Retrieve, Temporal Warp and Refine.

*1) Content Encoder $E_c$:* The Content Encoder $E_c$ (Fig. 3 (a)) aims at learning a content embedding that is dependent only on the speaking content. The main challenge here is to normalize the speaker-specific information. To achieve this goal, we extract Mel-Frequency Cepstral Coefficients (MFCC) features from the input audio waveform and process them by the Phonetic PosteriorGrams (PPG) network [48]. The PPG network is a Deep Bidirectional Long Short Term Memory based Recurrent Neural Network (DBLSTM) that produces PPG from input Mel-Frequency Cepstral Coefficients (MFCC) features of speech audio. From the input MFCC features, the PPG network learns a phoneme distribution map where the phonemes are only defined by words themselves. PPG is a time-versus-class matrix, which represents the posterior probabilities of each phonetic class for each specific time frame of one utterance. To further remove speaker information in the phoneme distribution, we find it beneficial to add an Instance Normalization (IN [49]) layer without affine transformation to the Content Encoder $E_c$. Formally, if the input of the IN layer is a feature map $M$ that contains $N_C$ channels and the dimension of each channel is $L$, then $M_c$ (the $c$-th channel of the input feature map) is normalized to $M'_c$ as follows:

$$M'_c = \frac{M_c - \mu_c}{\sigma_c}, \text{ where } c \in \{1, \cdots, N_C\} \quad (1)$$

where $\mu_c$ and $\sigma_c$ are the mean and standard variation of all elements of $M_c$, respectively. We also adopt Conv1d layers to handle the time-serial relevance of the produced phoneme distributions, and a ConvBank layer to capture the short-term and long-term speech information [50].

*2) Style Encoder $E_s$:* The Style Encoder $E_s$ (Fig. 3(b)) focuses on learning the speaker-specific speaking style embedding from the mouth motion, which is represented as a mouth landmark sequence. The unique speaking style contains the speaking timing and time-varying mouth shapes. Thus, we introduce the Conv1d and ConvBank layers to handle time-serial relevance that contains speaking timing information. The time-varying mouth shapes of a speaker are mainly captured by the global pattern of mouth movements so we introduce the average pooling layer. Thus, our Style Encoder can capture the unique speaking style of an unseen speaker.

*3) Cross-Modal AdaIN & Decoder $D$:* The Decoder $D$ (Fig. 3(c)) combines the content and the style embedding to infer the corresponding landmarks. To effectively fuse the speaking content (from audio) and style (from video), our cross-modal AdaIN [7] layer transfers the speaking style via low-order statistics (*i.e.*, mean and standard variation). Specifically, the Decoder $D$ normalizes the content information through an IN layer and the outputs are further modulated by the affine transformed style embedding. We use the same notations of the previous IN layer for the cross-modal AdaIN layer here. The cross-modal AdaIN layer performs the following operation:

$$M'_c = \gamma_c \frac{M_c - \mu_c}{\sigma_c} + \beta_c, \text{ where } c \in \{1, \cdots, N_D\} \quad (2)$$

where $\gamma_c$ and $\beta_c$ for each channel are the affine transformation parameters of the style embedding produced by the Style Encoder $E_s$.

The three modules $E_c$, $E_s$ and $D$ are trained jointly using paired audio clip $A_c$ (as the input source audio) and mouth landmark sequence $L_c$ (as the supervision target). The input target style $L_s$ is randomly sampled from all the mouth landmark sequences of the same speaker. The Style Translation Network reconstructs the mouth landmark sequence $L_c$ from speaking content of $A_c$ and speaking style of $L_s$. The training

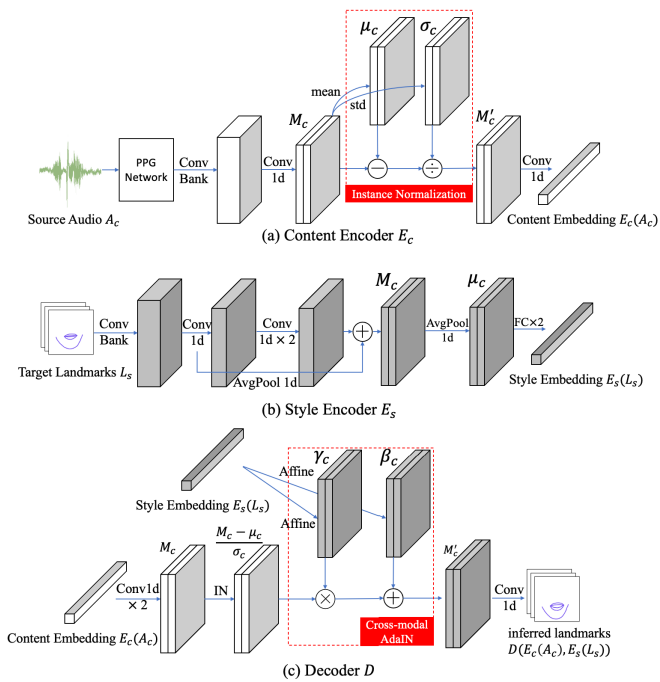

Fig. 3. **Style Translation Network for various speakers.** The Style Translation Network consists of Content Encoder $E_c$, Style Encoder $E_s$ and Decoder $D$. These components work together to transform the source audio to landmarks while preserving the unique style of the target speaker via the cross-modal AdaIN layer.

loss for the Style Translation Network is defined as:

$$L_{style} = ||D(E_c(A_c), E_s(L_s)) - L_c||_1 \quad (3)$$

### B. Retrieval-Based Video Renderer

The renderer is specifically devised for generating realistic and continuous videos even when the training data is limited. In UGC production, the limited video data make it hard to train a person-specific generator. Thus, we propose the retrieve-warp-refine pipeline, in which the existing mouth textures can be leveraged to facilitate texture generation. We first retrieve mouth textures and their landmarks by the inferred landmarks. Then, we propose the Temporal Warp module that warps retrieved mouth textures to make them match with the inferred landmarks and decreases the discontinuity between frames. Finally, we design a Refine module to further improve the video quality such as mouth interiors and fill the seams brought by retargeting warped mouths onto target faces.

*1) Retrieve:* For an unseen speaker, we first build a compact texture bank for the mouth area from its limited video data. We slide a fix-width time window (window size $N = 6$ in our experiment) on the mouth area and landmark sequence in the video. Inside the window, we can obtain a mouth image sequence $I_m^i$ and its corresponding mouth landmark sequence $L_m^i$, and they are paired as $(L_m^i, I_m^i)$. The mouth region of each frame is cropped and resized to the same size. Then, all the $N$ pairs constitute the texture bank $\{(L_m^i, I_m^i)\}_{i=1}^N$. During the inference phase, we retrieve the most similar mouth landmark and image sequence by comparing the difference between $L_m^i$ and $\hat{L}_c$ inferred by the Style Translation Network. Specifically, the most similar results are retrieved as follows:

$$s = \text{argmin}_i ||L_m^i - \hat{L}_c||_1, \text{ where } i \in \{1, \cdots, N\} \quad (4)$$

where $s$ represents the index of the retrieved mouth image landmark sequence pair. Note that we retrieve mouth images and landmarks in the form of continuous sequences, and such retrieved results are more continuous than the results composed of several single frames.

*2) Temporal Warp:* The Retrieve module returns mouth landmarks and images that best match the inferred landmarks. However, inevitable gaps still exist due to the limited size of the texture bank. To eliminate these gaps, we introduce the Temporal Warp module to warp the retrieved images to minimize the differences between the mouth movement of the retrieved landmarks and the inferred landmarks.

In particular, since the thin-plate spline (TPS) is effective in modeling the coordinate transformations, we apply TPS warping [9] on the retrieved mouth images. To make areas far from landmarks warped with high fidelity, we adopt radial basis functions in TPS to mainly warp areas around landmarks. In TPS, an interpolant function is also computed from the retrieved landmarks and inferred landmarks. Let $p_{i,t}$ and $v_{i,t}$ denote the $i$-th landmark coordinates of the retrieved and the inferred landmarks at time $t$, with $i \in \{1, \cdots, N\}$ and $t \in \{1, \cdots, T\}$. The mouth image size is $H \times W$. The TPS interpolant $f(x, y, t)$ minimizes the fitting error $E_f$ and the bending energy $E_b$ as follows:

$$E_f = \frac{1}{NT} \sum_{t=1}^{T} \sum_{i=1}^{N} ||f(p_{i,t}) - v_{i,t}||_1 \quad (5)$$

$$E_b = \frac{1}{WHT} \sum_{t=1}^{T} \sum_{x,y} f_{xx}^2(t) + 2f_{xy}^2(t) + f_{yy}^2(t) \quad (6)$$

where $f_{xy}$ means the partial derivatives of $f(x, y, t)$ for variable $x$, then for variable $y$. $f_{xx}$, $f_{yy}$ and $f_{tt}$ represent second-order partial derivatives of $f(x, y, t)$ for variable $x$, $y$ and $t$, respectively. The bending energy describes how "bending" the TPS interpolant $f(x, y, t)$ is. The TPS warp resists too much "bending" of the TPS interpolant since too much "bending" will cause stretch distortion. Thus, TPS warp implies a penalty involving the smoothness of the TPS interpolant $f(x, y, t)$. Here, we briefly explain the form of the bending energy. At a fixed time $t$, $f(x, y, t)$ is a matrix. The second-order partial derivatives of $f(x, y, t)$ for variable $x$ and $y$ regulate $f(x, y, t)$ to variate monotonously in the 2D plate space. In contrast to monotonous variation, the $f(x, y, t)$ will bend. For example, imagine that the left side of $f(x, y, t)$ moves right and its right side moves left, then the center part of $f(x, y, t)$ will bend. The second order derivate $f_{xx}$ in the bending energy will punish this bending. The naïve Thin Plate Spline (TPS) usually uses the radial basis function to represent a coordinate mapping from $\mathbb{R}^2$ to $\mathbb{R}^2$. We also apply the commonly used classical form in TPS warp [9]. Thus, the

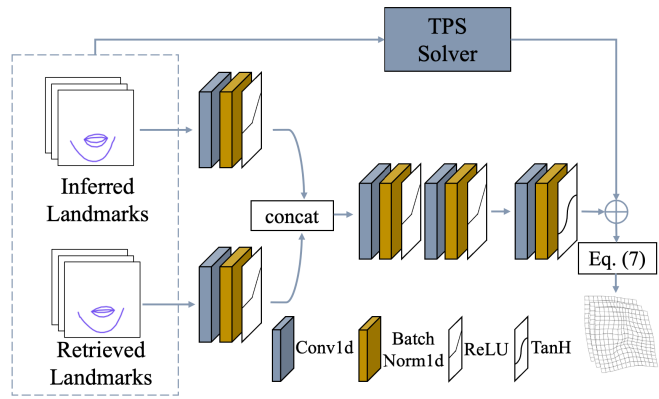

Fig. 4. **Temporal Warp.** Our Temporal Warp module extends TPS warp [9] to video level and considers video continuity.

TPS interpolant $f(x, y, t)$ has the form:

$$f(x, y, t) = a_{1,t} + a_{x,t}x + a_{y,t}y + \sum_{i=1}^{N} w_{i,t} U(||v_{i,t} - (x, y)||_1) \quad (7)$$

where $U(r) = r^2 \log r$, and $a_{1,t}, a_{x,t}, a_{y,t}, w_{i,t}$ can be solved by a linear system [9]. The above TPS interpolant will degrade to $f(x, y)$ for a single image (e.g. $T = 1$). In our method, the TPS interpolant $f$ maps the pixel at $(x, y)$ of the $t$-th retrieved mouth image to the new position $f(x, y, t)$ of the warped image.

The naïve TPS solver warps images frame by frame, neglecting the temporal continuity of the frame sequence. We reformulate the conventional TPS method by introducing a new temporal regularization $E_t$ that minimizes the second order derivative of the function $f(x, y, t)$ for time $t$ at a position $(x, y)$ on the whole plate. The minimization objective is as follows:

$$E_t = \frac{1}{WH} \sum_{x,y} f_{tt}^2(x, y) \quad (8)$$

The temporal regularization $E_t$ penalizes potential temporal jitter at each warping point. The Temporal Warp module optimizes the weighted sum of $E_f$, $E_b$ and $E_t$.

$$L_{tw} = \alpha_1 E_f + \alpha_2 E_b + \alpha_3 E_t \quad (9)$$

where $\alpha_1, \alpha_2$ and $\alpha_3$ are set empirically. We find that it is challenging to train a network from scratch to optimize $L_{tw}$. To tackle this problem, we design the Temporal Warp module as shown in Fig. 4. We train a network to learn the parameters of the TPS interpolant function based on the solution of the naïve TPS. Specifically, the network is trained to learn the residual of the solution of the existing TPS solver. In Fig. 4, the TPS solver outputs the parameters $a_{1,t}$, $a_{x,t}$, $a_{y,t}$ and $w_{i,t}$ in Eq. (7) solved by the algorithm in [9]. In Fig. 4, the network outputs residuals for the solved parameters $a_{1,t}$, $a_{x,t}$, $a_{y,t}$ and $w_{i,t}$. As shown in Fig. 4, our Temporal Warp module produces a "bending thin sheet". The bending energy in Eq. (6) describes how "bending" the "thin sheet" is. The module minimizes $L_{tw}$ and outputs $a_{1,t}, a_{x,t}, a_{y,t}, w_{i,t}$. The final interpolant is calculated by Eq. (7).

*3) Refine:* Seams might emerge in the results of retargeting the warped mouth images to the target face images, as shown in Fig. 2. We propose the Refine module to eliminate these seams as well as the minor head pose and illumination changes between target frames and retrieved mouth regions. This module also improves details of the generated sequence. We provide more details as follows. We first retarget the warped mouth by the mouth center computed from the mouth landmarks. The face image sequence with warped mouths $I_w$ is fed to our U-Net based [51] Refine module $R$ to generate seamless sequence $I_r$. Let $I_w[t], I_r[t]$ denote the $t$-th image in image sequence $I_w$, $I_r$ respectively, where $t \in \{1, \cdots, T\}$. During the training phase, we randomly sample mouth landmark and image sequence pair $(L_{gt}, I_{gt})$, where $L_{gt}$ is regarded as the set of inferred landmarks and $I_{gt}$ plays the role of supervision for the Refine module.

We train the Temporal Warp and Refine module jointly. We introduce the following objective to optimize these modules.

$$L_r = \beta_1 L_{rec} + \beta_2 L_{frame} + \beta_3 L_{seq} + \beta_4 L_{vgg} + \beta_5 L_{tv} \quad (10)$$

where $L_{rec}$ is a pixel-wise L1 loss between the generated sequence $I_r$ and the ground truth $I_{gt}$. $L_{rec}$ accelerates the optimization but it overly smooths the generated results. The idea of GAN [52], [53] is used to improve the realism of generated images. In particular, to generate photo-realistic face images and temporally coherent face videos, we introduce an Image Discriminator [54] and a Video Discriminator [54] in our method to improve image realism and video continuity, respectively. The Frame Discriminator $D_f$ maximizes $L_{frame}$ while the Refine module $R$ minimizes $L_{frame}$ as follows:

$$L_{frame} = \frac{1}{T} \sum_{t=1}^{T} \log D_f(I_{gt}[t]) - \log D_f(R(I_w)[t]) \quad (11)$$

Similarly, the Sequence Discriminator $D_s$ competes with the Refine module $R$ to improve the continuity of generated videos by alternatively maximizing and minimizing $L_{seq}$ as follows:

$$L_{seq} = \log D_s(I_{gt}) - \log D_s(R(I_w)) \quad (12)$$

In addition, we also use the perceptual loss $L_{vgg}$ [55] to improve the generated image quality by constraining the image features at different scales, and using a standard Total Variation loss $L_{tv}$ to reduce spiky artifacts brought by $L_{vgg}$ [55].

After refining the mouth textures by the Refine module, we use Laplacian blending to compose the refined mouth texture and the background target face. We first form a 0-1 mask (the black and white mask in Fig. 2) by connecting the uppermost point and the jawline points of the inferred landmarks. Then, the mouth texture is selected by the white area and the face background is selected by the black area. The refined mouth texture and the face background of the target video are composed by the Laplacian blending algorithm.

### C. Training Details

To train the Style Translation Network, we prepare the inputs as follows: i) the source audio $A_c$ that provides speaking content and ii) the target landmark sequence $L_s$ that provides the speaking style. The learning target $L_c$ is the landmark sequence extracted from the video clip corresponding to the input $A_c$. The input $L_s$ is randomly sampled from the texture bank of the speaker, of whom we obtain $A_c$. To train the Retrieval-based Video Renderer, we randomly sample two pairs of landmark and image sequence $(L_{gt}, I_{gt})$ and $(L_{re}, I_{re})$, where $(L_{re}, I_{re})$ is used as the retrieved result and $L_{gt}$, $I_{gt}$ is used as the target inferred landmarks and mouth textures. Since $(L_{re}, I_{re})$ is randomly sampled during training, the trained Temporal Warp module is capable of handling harder warping and refining problems than those during testing. In our method, we apply a sliding window of 6 frames to capture all the input and ground truth sequences.

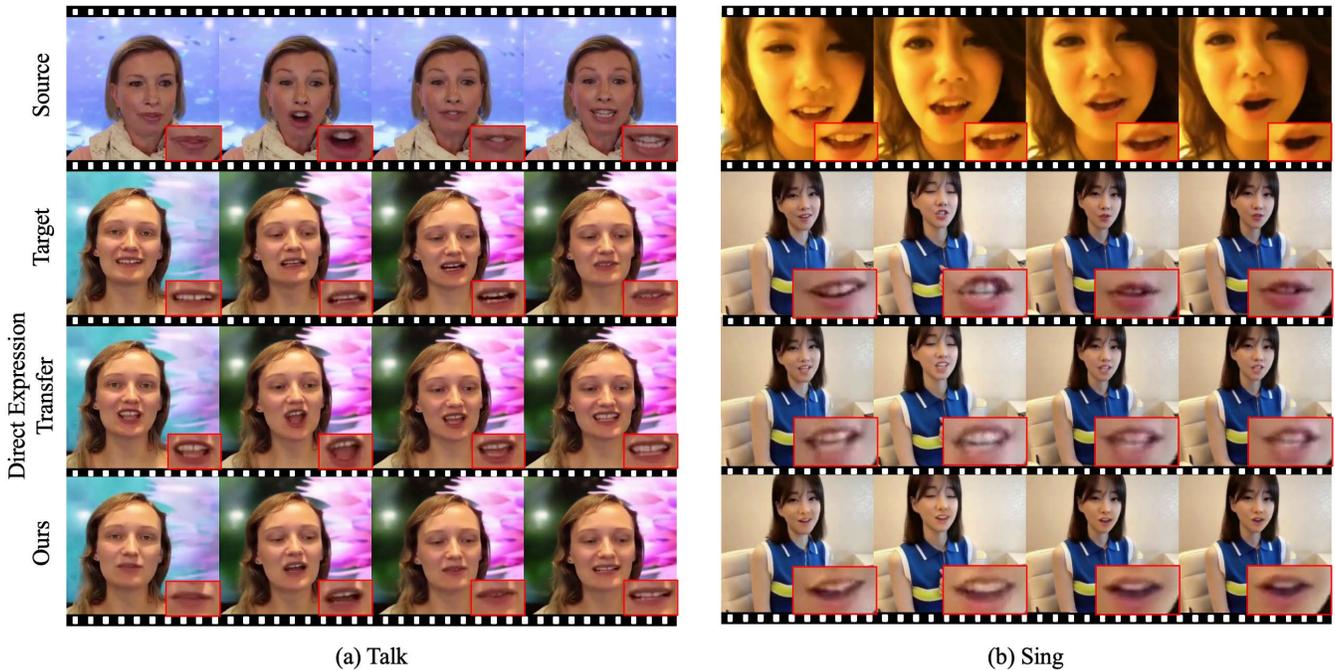

Fig. 5. **(a) Other-Drive Results of Talking.** Our model generates a style preserved talking video by the audio of another speaker. **(b) Other-Drive Results of Singing.** Our model generates a style preserved singing video by the audio of another speaker.

At first, we separately pre-train the Style Translation Network and Retrieval-based Video Renderer on the RAVDESS [56] dataset. Then, for any new speaker in hand, we finetune our whole model on its short video clip. The pre-training phase decreases the finetuning time of our parametric Refine module. Recent dubbing methods either generate mouth texture from scratch [27] or synthesize face videos based on rendered mouth texture [34], [36]. Thus, these methods benefit less from the pre-train & finetune strategy. The design of the retrieve-warp-refine pipeline in our Retrieval-based Video Renderer reduces the difficulty of transferring our model to any new speaker. Thus, our method can be finetuned with just a handful of data of a new speaker efficiently yet producing high-quality retargeted face videos.

In our network optimization, we set $\alpha_1 = \alpha_2 = \alpha_3 = 1$ in $L_{tw}$ since the three functions play equal role in optimizing the Temporal Warp module. We balance the influence of loss functions in $L_r$ with $\beta_1 = 10$, $\beta_2 = 1$, $\beta_3 = 0.1$, $\beta_4 = 1$ and $\beta_5 = 1$. In the experiment, we find that the high weight $\beta_1 = 10$ of $L_{rec}$ accelerates the optimization of $L_{rec}$. We set a small weight $\beta_3 = 0.1$ on $L_{seq}$ to ensure the temporal continuity improved by the Temporal Warp module and make the Refine module focus on optimization for the visual quality of frames. The loss $L_{style}$ and $L_r$ are minimized by the Adam solver [57] with an initial learning rate of $10^{-3}$ and an exponential decay rate of 0.5. Also, we use the Adam solver with an exponential decay rate of 0.5 to minimize the loss $L_{tw}$ but we empirically find that the training of the Temporal Warp module only converges when the learning rate is as small as $10^{-6}$. The latter learning rate is much smaller than the former one, hence, we firstly train the Temporal Warp module alone, then train the whole Retrieval-based Video Renderer with parameters of the Temporal Warp module frozen. To avoid exploding gradients, we apply the gradient norm clipping operation [58] during the back-propagation for our network.

## IV. EXPERIMENTS

We present the capability of our audio dubbing method for UGC production in Sec. IV-A. We compare our approach with recent state-of-the-art methods designed for PGC production in Sec. IV-B. Then, we compare the training time, training data requirement and testing speed of our method with recent audio dubbing methods in Sec. IV-C. We also perform an ablation study to validate the effects of our proposed components in Sec. IV-D. Finally, we conduct web-based user studies to validate the superiority of our method over recent methods and the effects of the proposed components.

*1) Dataset:* We evaluate our model on benchmark talking face dataset RAVDESS [56] and online video clips from

the Internet. The RAVDESS dataset contains 7, 356 talking video clips by 12 male and 12 female speakers, where each speaker sings/speaks the same content in different emotions twice. We use the data of the first 20 speakers in the RAVDESS dataset to pretrain our Style Translation Network and Retrieval-based Video Renderer. For each unseen testing speaker, the video length is *no more than* 30 *seconds* and the information of these videos can be viewed in the *supplementary material*. For each result in our paper and *supplementary video*, we also list the detailed settings about training data, testing source audio and testing style video in the *supplementary material*.

*2) Preprocessing:* We re-sample the audio and video so that the audio sample rate is $44.1kHz$ and the video fps is 30. For the audio signal, we calculate the MFCCs (Mel-Frequency Cepstral Coefficients) feature on the sliding window of $200ms$ and remain the first 20-D features. The final shape of the audio feature is $120 \times 40$, where 120 is the time dimension corresponding to consecutive $N = 6$ video frames, and 40 is the feature dimension that contains 20-D MFCCs and 20-D temporal derivatives. For the video signal, we extract all the frames from portrait videos and detect the 106 facial landmarks by [59]. We coarsely align the human face images according to the 5 landmarks: the center of the left eye, the center of the right eye, the nose tip, and the left and right corners of the mouth. From these 5 landmarks, we align the face image by the affine transform algorithm to eliminate the problem of head locations and unfixed heights. The 20 lip landmarks and 19 jawline landmarks are defined as the 39 "mouth area" landmarks in our method. We define the average of 20 lip landmarks as the mouth center $(x_0, y_0)$ and crop a $148 \times 148$ square mouth area (left-top:$(x_0 - 74, y_0 - 74)$, right-bottom:$(x_0 + 74, y_0 + 74)$) as mouth texture to store in the Texture Bank.

*3) Metrics:* We evaluate the quality of the generated talking videos in the visual quality, lip-sync accuracy and naturalness. The naturalness of generated videos refers to what degree that the head motion looks natural and matches the speaking content. For quantitative metrics, we use FID (Fréchet Inception Distance [60]) to evaluate the video frame visual quality and SyncNet Dist. (audio and mouth shape distance calculated by SyncNet [61]) to evaluate the lip-sync accuracy. We use a per-pixel distance map and mean photometric error of mouth area to measure the continuity of generated videos. For video naturalness, continuity and style-preserving degree that are hard to measure in objective metrics, we use user studies to evaluate them. We also conduct A-B compare user studies to compare our method with recent methods and validate the effects of our proposed modules.

To quantitatively evaluate how well the Style Encoder captures speaking styles. We propose a new metric called Speaking Style Intersection Over Union (SSIOU). First, since speaking style differences include variation in lip articulation, we build the distance between the lower and upper lip as a function of time. Then, for the same speaking content, the area under the function curve (RUC) is used to describe the speaking styles of different speakers. Thus, for the same speaking content, we use the IOU of the RUCs of different speakers to measure the similarity of their speaking styles. We call it as Speaking Style Intersection Over Union (SSIOU).

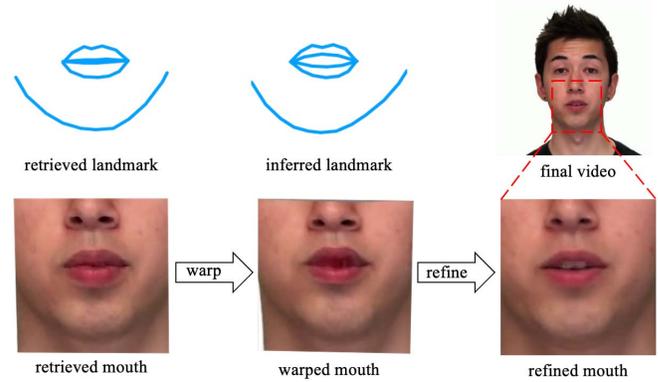

Fig. 6. **Hard Case.** We present a hard case that the retrieved landmarks are largely different from the inferred landmarks. Benefiting from the generation capability of the Refine module, the missing texture inside the mouth is well generated.

### A. Audio-Driven Dubbing

*1) Self-Driven Results:* After training our model on the unseen speaker's video clip of 30 seconds for 25 minutes, our model generates photo-realistic and audio-aligned talking videos as shown in Fig. 7 and the *supplementary video*. Note that our audio dubbing method requires a low amount of training data and training time, which will open a broader range for UGC applications. In addition, our model can generate the talking video at 120ms per frame, which is faster than most dubbing methods designed for PGC production (please refer to the *supplementary material*).

*2) Other-Driven Results:* Previous audio dubbing methods [6], [27] pay little attention to the speaking style of the target speaker. Thus, when the source audio does not belong to the target speaker, the inconsistency between the speaking styles of generated and true videos might be detected. The generated talking and singing results are shown in Fig. 5 and the *supplementary video*. Our method learns the mouth movement from the source audio and preserves its speaking style from the video of the target speaker.

*3) Hard Case and Analysis:* For most cases, the limited data used to build the texture bank contains mouths of different distances between the upper lip and the lower lip. Thus, the given limited video footage is enough to cover the mouth movements of different visemes. However, landmarks in the texture bank might be largely different from the inferred landmarks. For example, the landmarks in the texture bank only contain closed mouths while the inferred landmarks are wide-open mouths. We demonstrate the detailed intermediate results like the retrieved landmark and mouth, inferred landmark, warped mouth and refined mouth in Fig. 6 for these hard cases. From Fig. 6, we can see that the Temporal Warp module transforms the retrieved mouth textures even if the texture inside the mouth is missing. Thanks to the generation capability of the Refine module, the missing texture inside the mouth is well generated in the final video. As shown in

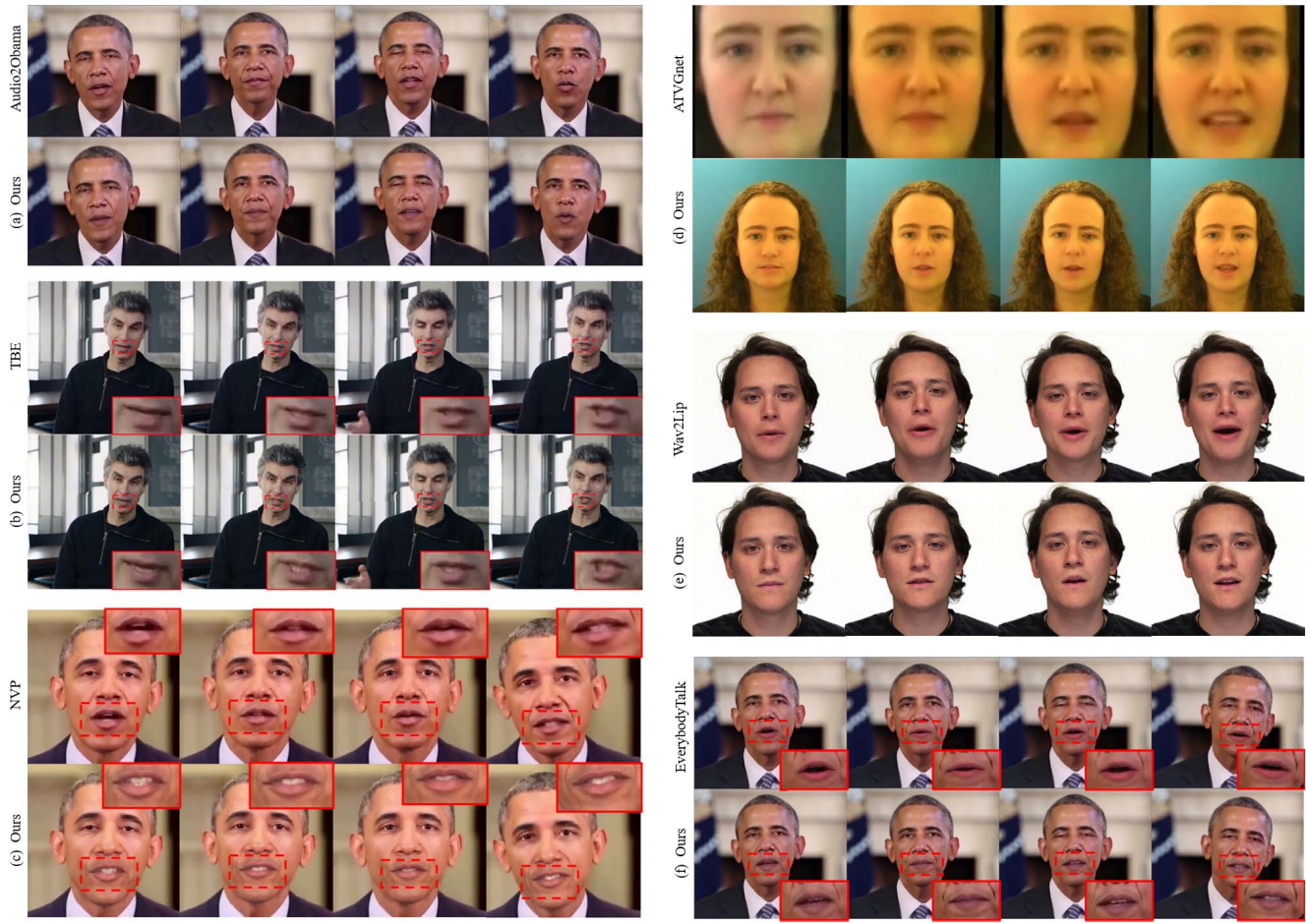

Fig. 7. **Comparison with Audio/Text driven Dubbing Methods.** Compared to Audio2Obama, TBE and NVP, our model generates competitive results with a smaller training data size. Compared with EverybodyTalk, our model generates high quality videos, especially details like teeth.

Fig. 6, our proposed method can be well generalized to cases where the inferred landmarks are largely different from the landmarks in the texture bank.

### B. Comparisons With the State-of-the-Art Methods

We conduct extensive experiments to compare our method with recent audio/text driven dubbing methods, including Audio2Obama [5], TBE [25], NVP [6] and EverbodyTalk [27]. We also compare our method with visual dubbing methods like Face2Face [33], DVP [34], and Style-preserving VDub [36]. We recommend viewing the *supplementary video* for more details.

*1) Comparisons to Audio/Text Driven Dubbing Methods:* We compare the visual quality (FID) and lip-sync accuracy (SyncNet Dist.) of our results and those of recent audio/text-driven dubbing methods in Tab. I. Compared with Audio2Obama, our method achieves very competitive results in audio-visual alignment and mouth details like teeth and nasolabial folds. The comparison details can be viewed in Fig. 7 (a). Note that our method requires 30 seconds of training data of Obama while Audio2Obama requires up to 14 hours of training data. Then, compared with the retrieval-based text-driven method TBE, our method also produces competitive talking videos. From Fig. 7 (b), we can see that our method generates competitive textures of teeth. In addition, the viseme search and retiming in TBE might break the speaking style lying in consecutive mouth frames and it takes more computing resources (training 42 hours on 1 hour video data). In addition, we compare our method with NVP in Fig. 7 (c). Our method achieves competitive results. For a new speaker, the NVP needs to train more than 30 hours on a 2-3 minutes video. The training time requirement limits its application in UGC production. Also, we compare our method with ATVGnet [62] that generates talking videos from audio and a still face image in Fig. 7 (d). Our method can produce more realistic talking videos since our retrieve-warp-refine pipeline also considers the background in the generated videos. At last, we compare our method with the recent popular Wav2Lip [63] in Fig. 7 (e). This method is designed for large-scale datasets and performs not well in the UGC production as the authors also claim that fine-tuning Wav2Lip on a few minutes of a single speaker might not get good results [63]. Thus, our method performs better than Wav2Lip in the UGC production.

We also compare our method with EverybodyTalk [27] in Fig. 7 (f). We use our training strategy (pretraining on the RAVDESS dataset and training on a 30-second talking video of the new speaker) to train the network of EverybodyTalk.

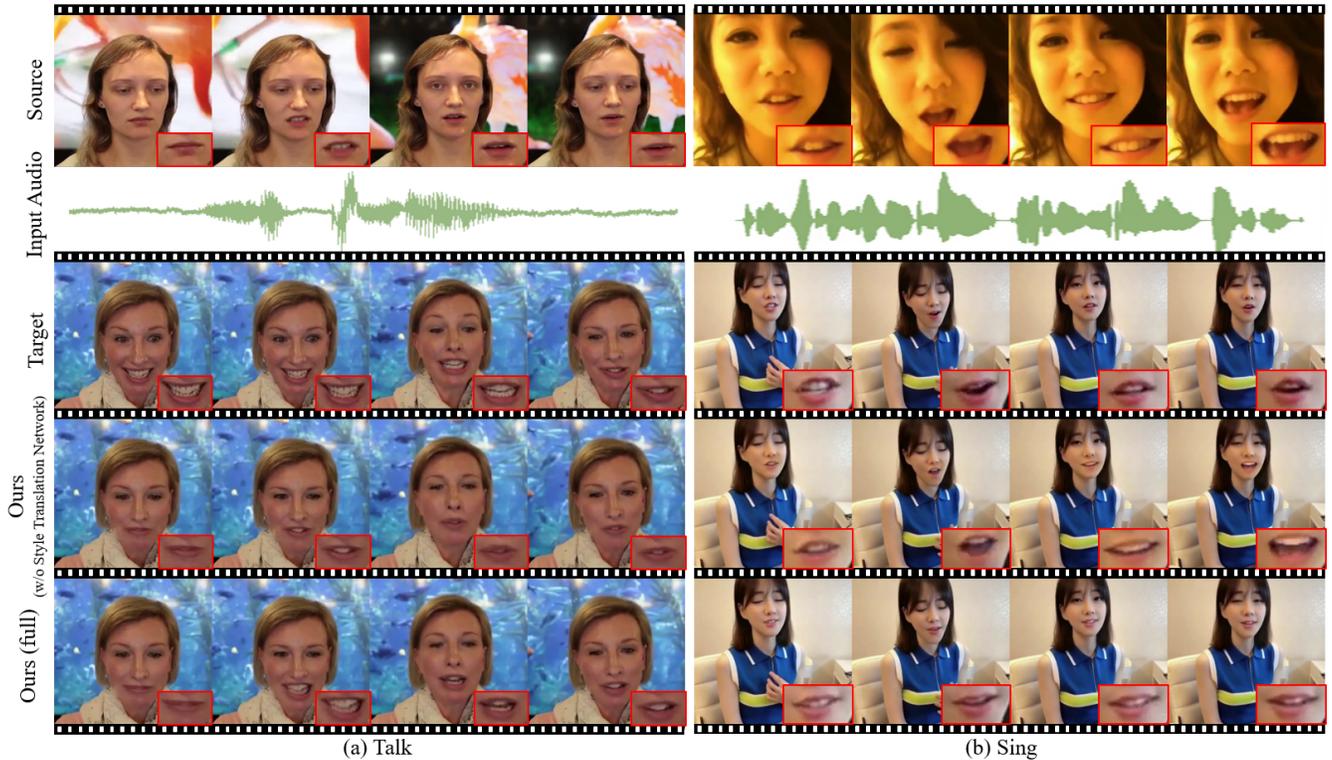

Fig. 8. **Effect of the Style Translation Network.** We compare the (a) talking / (b) singing videos generated by our model without style translation and our full model. Note that different speakers open their mouths at different scales. Our full model better preserves the speaking style of the target actor, which is best viewed in the *supplementary video*.

TABLE I
COMPARISON OF IMAGE QUALITY AND LIP-SYNC ACCURACY. WE COMPARE OUR METHOD WITH RECENT STATE-OF-THE-ART AUDIO-DRIVEN DUBBING AND VISUAL DUBBING METHODS IN IMAGE QUALITY (FID) AND LIP-SYNC ACCURACY (SYNCNET DIST). WE ALSO LIST THE ABLATION STUDY RESULTS IN THIS TABLE

| Method | FID | SyncNet Dist. |
| --- | --- | --- |
| Audio2Obama/Ours | **0.656**/0.659 | **7.96**/10.78 |
| TBE/Ours | 0.25/**0.20** | 10.39/**8.80** |
| NVP/Ours | 0.72/**0.65** | 10.89/**10.15** |
| ATVGnet/Ours | 0.84/**0.60** | 11.59/**9.87** |
| Wav2Lip/Ours | 0.70/**0.59** | 11.73/**9.94** |
| EverybodyTalk/Ours | 0.23/**0.13** | 11.53/**10.67** |
| Face2Face/Ours | 0.72/**0.65** | 11.04/**10.59** |
| DVP/Ours | 0.80/**0.20** | 10.92/**8.80** |
| Single frame/Ours | 0.60/**0.49** | 10.44/**9.21** |
| Fixed sequence | 0.57/**0.49** | 10.12/**9.21** |
| w/o Style Translation/Ours | 0.62/**0.49** | 10.92/**9.21** |
| w/o Temporal Warp/Ours | 0.54/**0.49** | 10.46/**9.21** |
| w/o Refine/Ours | 0.59/**0.49** | 10.80/**9.21** |

EverybodyTalk relies on a time-consuming teeth proxy (300ms per frame) to inpaint teeth texture, which might fail in retrieving matched teeth texture when the training data is very limited. Thus, the result produced by EverybodyTalk lacks mouth details like teeth while our method generates audio-visual aligned video with better mouth details. From Tab. I, we can see that our method outperforms TBE, NVP, ATVGnet, Wav2Lip and EverybodyTalk. Our method also achieves very competitive results compared with Audio2Obama even though it is trained on 14h video data of Obama while our method only requires 30s.

*2) Comparisons to Visual Dubbing Methods:* Visual Dubbing methods directly transfer mouth movement from the source actor to the target actor. However, visual dubbing methods rely on the capture of face movements from source videos, which narrows their application range. Our method tries to solve a more challenging problem: learning mouth movement directly from the source audio without any visual cues. Though our audio-driven method lacks visual cues, it still performs comparably to the state-of-the-art visual dubbing methods. We present the quantitative comparison results in Tab. I.

First, we compare to Face2Face [33] in Fig. 9 (a). The Face2Face [33] transfers facial expressions inferred by 3D face morphable model [64], [65], [66], retrieves mouth texture by the facial expression and renders the talking face. Our method produces competitive results with realistic texture and mouth movement, suggesting that our Style Translation Network learns accurate mouth movement from the input audio. Besides, the head motion in our result is more natural than that of Face2Face. Then, we compare to DVP [34] in Fig. 9 (b). The DVP method requires time-consuming face rendering and its Rendering-to-Video Translation Network might fail in generating complex background. Our method generates better talking face videos with a smaller training data size, shorter training time and faster inference speed. As presented in Tab. I, our method achieves better FID and SyncNet Dist. thanks to the retrieval-based pipeline and Style Translation Network. Finally, we compare to Style-Preserving VDub [36] in Fig. 10. Our method achieves competitive results. Our audio-driven method learns accurate mouth movement from audio and preserves the speaking style of the target actor. In addition, our method requires a smaller training data size and has a faster inference speed.

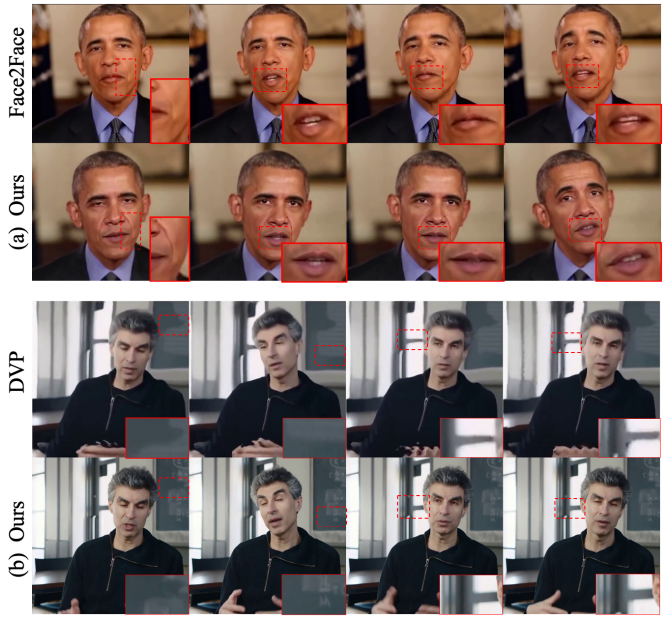

Fig. 9. **Comparison to Visual Dubbing Methods.** The similar lip movements validate that our Style Translation Network learns accurate lip movements from input audios. Our Retrieval-based Video Renderer also generates competitive video details.

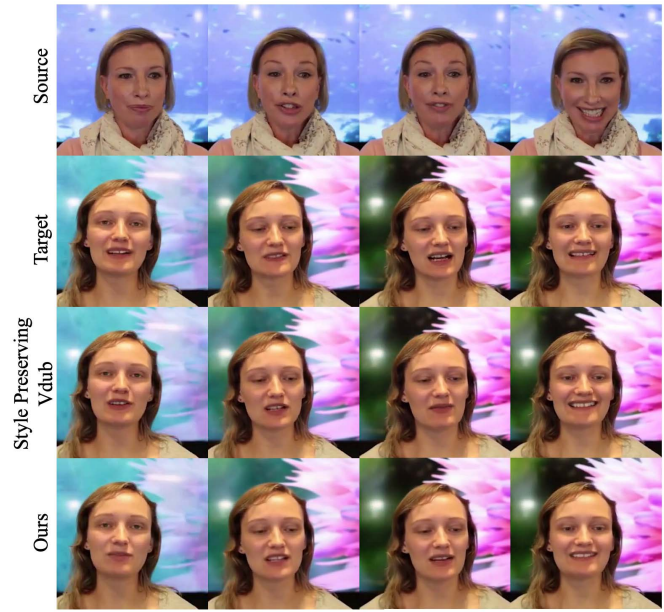

Fig. 10. **Comparison to Style-Preserving VDub.** Our model learns the mouth movement from source audio. Moreover, our model also preserves the speaking style of the target actor, like the Style-Preserving VDub [36] method. Our method tries to solve an ill-posed problem: learning mouth movement representation from source audio, while the Style-Preserving VDub [36] method transfers mouth movement representation from the source video. Even though, our method still generates competitive results in mouth details and style preservation.

TABLE II
TRAINING AND RUNTIME PERFORMANCE COMPARISON. WE COMPARE OUR METHOD WITH RECENT AUDIO-DRIVEN DUBBING METHODS ON TRAINING DATA SIZE, TRAINING TIME AND TESTING SPEED

| Method | Training | | Testing |
|---|---|---|---|
| | Training Time | Data Size | Time/Frame |
| Audio2Obama | 2 hours | up to 14 hours | 1.5s |
| TBE | 42 hours | 1 hour | >132ms |
| NVP(GTX 1080Ti) | >30 hours | 2-3 minutes | **real time** |
| EverybodyTalk | 8 hours | 15 minutes | 1.7s |
| Ours | **25 minutes** | **30s** | 120ms |

### C. Computing Resource and Inference Speed Comparison

We compare our method to recent audio-driven dubbing methods in training data size, training time and testing speed. Specifically, we use the same NVIDIA TitanX GPU to evaluate the training and testing time for Audio2Obama, TBE, EverybodyTalk, NVP and our method.

Tab. II lists the training time, training data requirement and testing speed. Our method requires a low amount of training data and training time. Benefiting from our training strategy and retrieve-warp-refine pipeline of the Retrieval-based Video Renderer, our method only needs to be finetuned on the new data for 25 minutes to generate talking face videos for a new speaker. While Audio2Obama, TBE, EverybodyTalk and NVP have to retrain the speaker-specific rendering network on the new data, which will take much longer time.

The testing speed of NVP is faster than our method. Since NVP takes much more training time (30h v.s. 25min), our method is better than NVP in User Generated Content (UGC) production illustrated as follows: For an unseen speaker, NVP [6] requires training more than 30 hours and its inference speed is about 9ms per frame (Sec. *Inference* in [6]). Our method needs to train for 25 minutes and the inference speed is about 120ms per frame. Here we assume that NVP and our method take the same time to generate talking videos of $x$ frames. Thus:

$$30 \times 3600 \times 1000 + 9 \times x = 25 \times 60 \times 1000 + 120 \times x \quad (13)$$

where the left and right sides are the time consumed by NVP and our method, respectively. The solution is $x \approx 959459$. If the video fps is 25, then the video length of $x$ frames is 10.7 hours. Thus, our method takes less time than NVP for generating talking videos if the generated video length is less than 10.7 hours. In consideration of time consumption, our method is a much better choice for UGC production.

### D. Ablation Study

*1) Effect of the Style Translation Network:* To validate that the Style Translation Network preserves the speaking style of the target speaker. We train a new model that modifies the Style Translation Network. Specifically, we remove the Style Encoder and the cross-modal AdaIN [7] module in the Decoder. Then, the Style Translation Network degrades into an Audio-to-landmark Mapping Network that no style information is considered. The visual results are presented in

TABLE III
QUANTITATIVE COMPARISON OF STYLE TRANSLATION NETWORK.
WE COMPARE THE SPEAKING STYLE SIMILARITY BY SSIOU
WITH/WITHOUT THE STYLE TRANSLATION NETWORK OF 4 SPEAKERS

| Speaker ID | 1 | 2 | 3 | 4 |
|---|---|---|---|---|
| w/ Style Translation Network | 0.82 | 0.72 | 0.78 | 0.84 |
| w/o Style Translation Network | 0.64 | 0.61 | 0.59 | 0.61 |

TABLE IV
QUANTITATIVE COMPARISON OF TPS AND TEMPORAL WARP
MODULE. WE COMPARE OUR TEMPORAL WARP MODULE
WITH TPS WARP ON $E_f$, $E_b$ AND $E_t$

| Method | No Warp | TPS Warp | Temporal Warp |
|---|---|---|---|
| $E_f$ | 3.35 | 1.12 | **0.92** |
| $E_b$ | - | 0.0286 | **0.0286** |
| $E_t$ | - | 0.48 | **0.44** |

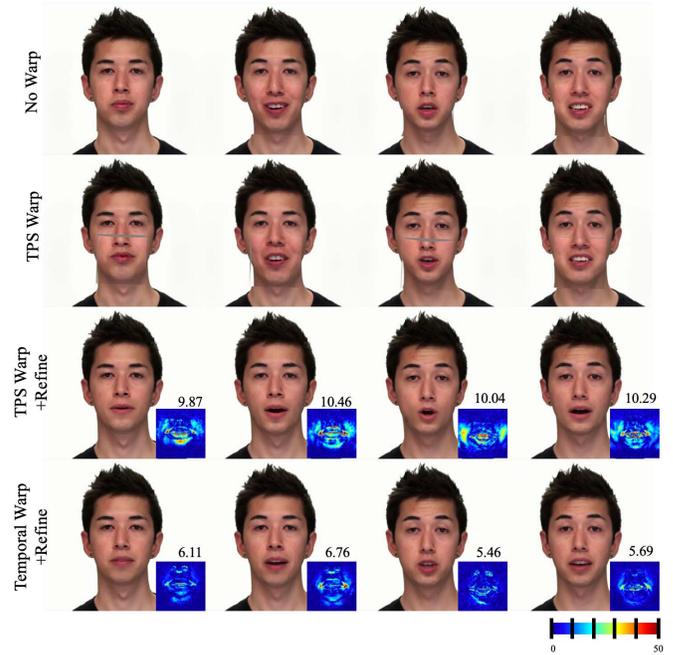

Fig. 11. **Qualitative Comparison of TPS and Temporal Warp Module.** Our Temporal Warp module generates talking videos with better temporal continuity than the TPS warp, best viewed in the *supplementary video*. Compared with TPS Warp, our Temporal Warp module performs better in per-pixel distance map and mean photometric error of generated mouth area. Thus, the Temporal Warp module promotes significant temporal stability (best viewed in the *supplementary video*). "No Warp" means directly using the retrieved mouth images.

Fig. 8 and the *supplementary video*. The 3rd row in Fig. 8 presents that the Audio-to-landmark Mapping Network simply retains the speaking style of the source actor. Our Style Translation Network can preserve the unique speaking style as shown in the 4th row in Fig. 8. It also supports the transfer of singing style between actors (*supplementary video*). We quantitatively compare visual quality and lip-sync accuracy in Tab. I and conduct a user study in Sec. IV-E to validate the effect of the Style Translation Network.

To further demonstrate the effect of the Style Translation Network, especially the ability of the Style Encoder in capturing the speaking styles, we compare SSIOU of different speakers before and after applying the Style Translation Network. First, we use 4 testing speakers (2 males and 2 females) from the RAVDESS dataset and group their video clips by the speaking content. Then, for the same speaking content, we calculate the SSIOU between the generated video and the ground truth video with/without the proposed Style Translation Network. We present the mean SSIOU results in Tab. III. Compared to our method without the Style Translation Network, our full method can better preserve the speaking style captured from the input landmarks.

*2) Effect of the Temporal Warp and Refine Module:* We extend the image-level TPS warp algorithm to our video-level Temporal Warp module to improve the continuity of the generated video. The quantitative comparison between TPS and our Temporal Warp module is presented in Tab. IV, in which our Temporal Warp module lower not only $E_t$, but also $E_f$. The decrease of $E_f$ can be explained that our video-level Temporal Warp module considers more global information, which helps the image-level warping. The qualitative comparison is presented in Fig. 11 and the *supplementary video*, where the Refine module inpaints the seams and improves visual quality. The mouth areas of the four videos are from direct retrieve (no warp), warping the retrieved mouth by TPS, refining the warping result by TPS, and refining the warping results by Temporal Warp module, respectively. In the last two rows of Fig. 11, we demonstrate the generated mouth area per-pixel distance map in the bottom-right corner, and the mean photometric error is shown above the error map. The averaged mean photometric error in the testing set of the RAVDESS dataset is decreased from **9.93** to **6.60** after replacing the TPS Warp with the proposed Temporal Warp module. In Fig. 11 and *supplementary video*, the accuracy of lip synchronization and the video temporal continuity are gradually improved.

The Refine module, formed as an UNet-based GAN, can generate missing pixels in the final video. As shown in Fig. 6, the Refine module generates the missing texture inside the mouth caused by stretching the closed lips. As shown in Fig. 11, the Refine module generates the texture inside seams between the warped mouth and the background face. Thus, supervised by the Frame Discriminator and the Sequence Discriminator, the Refine module helps to generate realistic and continuous video frames. We quantitatively compare the visual quality and lip-sync accuracy in Tab. I and conduct a user study in Sec. IV-E to validate the temporal continuity improvement brought by our Temporal Warp module and Refine module.

*3) Effect of Randomly Sampled Target Landmarks:* In the Style Translation Network, the target landmarks extracted from the target video are used to extract speaking style information by our proposed Style Encoder module. In practice, we randomly sample the mouth landmarks of the target speaker for the Style Encoder to learn robust speaking style embeddings. We also try to simplify this step to replace the randomly sampled mouth landmarks as a fixed landmark sequence or even the landmark of a single frame. We present the result in Tab. I. We compare our full method with our method that replaces the randomly sampled target landmark with a fixed landmark sequence in the row "Fixed sequence/Ours". We compare our full method with our method that replaces the randomly sampled target landmark with the landmark of a single frame in the row "Single frame/Ours". From the comparison results, we can see that our full method performs better than our method that uses a fixed landmark sequence and our method with the landmark of a single frame. Since speaking style is a dynamic feature that should be realized from a consecutive video, our method with the landmark of a single frame does not perform well. In addition, from a fixed landmark sequence, it is hard to learn a robust speaking style. Thus, our full method that randomly samples target landmarks can learn better speaking style features. We also conduct a user study to validate that our full method performs better in preserving speaking styles of target speakers in Sec. IV-E.

*E. User Study*

To quantitatively evaluate the effects of our proposed modules (Style Translation Network, Temporal Warp module and Refine module) and the superiority of our method over recent state-of-the-art methods, we conduct a web-based user study with 100 participants. Specifically, we collect generated videos of unseen speakers to evaluate the effect of the Style Translation Network, the Temporal Warp module, the Refine module, and compare our method to recent state-of-art audio-driven dubbing methods.

*1) User Study on Style Translation Network:* In each testing case, we present four videos to each participant. The four videos include two true videos that provide the source audio and the target speaking style, and two fake videos are generated by the Audio-to-landmark Mapping Network and the Style Translation Network. At first, the two true videos are shown to the participant. Then, we present the two generated videos and ask the participant which generated video better preserves the speaking style. Our user study contains 4 speaking and 4 singing videos of 4 speakers, and the number of testing cases is $4 \times (4 + 4) = 32$. The user study result is shown in Tab. V. The Style Translation Network is considered as better preserve the speaking style of the target actor in 78% of all cases.

*2) User Study on Temporal Warp Network and Refine Module:* We collect 32 pairs of generated videos where each video pair is generated by TPS + Refine or Temporal Warp + Refine from the same input data. The 32 pairs are from 4 speakers and each speaker provides 4 speaking and 4 singing clips. We present the pair to each participant and ask which video is more continuous. The study result is shown in Tab. V and our proposed modules are regarded as effective in most cases. Our Temporal Warp module is considered as more continuous in 65% of all cases. Similarly, we also collect 32 pairs of generated videos where each video pair is generated by Temporal Warp or Temporal Warp + Refine from the same input data. We present the pair to each participant and ask which video is better in visual quality. Our Refine module is considered to improve the video quality in 65% of all cases.

*3) User Study on Randomly Sampled Target Landmarks:* We collect two sets of generated videos. In the first set,

TABLE V
USER STUDY: ABLATION STUDY. WE CONDUCT WEB-BASED USER STUDIES ON THE EFFECT OF THE STYLE TRANSLATION NETWORK AND THE TEMPORAL WARP MODULE

| Methods | Preferred Rate |
|---|---|
| Ours > Single frame | 70% |
| Ours > Fixed sequence | 64% |
| Ours > w/o Style Translation | 78% |
| Ours > w/o Temporal Warp | 65% |
| Ours > w/o Refine | 65% |

TABLE VI
USER STUDY: COMPARISON. WE CONDUCT A WEB-BASED USER STUDY TO COMPARE OUR METHOD AND RECENT AUDIO/TEXT BASED METHODS. WITH SIGNIFICANTLY SHORTER TRAINING TIME AND LESS TRAINING DATA, OUR METHOD STILL ACHIEVES COMPETITIVE VIDEO RESULTS

| Methods | Visual quality | Lip-sync | Naturalness |
|---|---|---|---|
| Ours > Audio2Obama | 46% | 43% | 42% |
| Ours > TBE | 50% | 51% | 52% |
| Ours > NVP | 46% | 43% | 43% |
| Ours > ATVGnet | 76% | 70% | 82% |
| Ours > Wav2Lip | 70% | 64% | 47% |
| Ours > EverybodyTalk | 65% | 62% | 67% |
| Ours > Face2Face | 74% | 72% | 82% |
| Ours > DVP | 96% | 97% | 86% |

we compare our full method with our method that replaces the randomly sampled target landmark with the landmark of a single frame. In each testing case, we present four videos to each participant, including two true videos that provide the source audio and the target speaking style, and two videos generated by our full method and our method that uses the landmark of a single frame. In the second set, we compare our full method with our method that replaces randomly sampled target landmarks with the fixed landmark sequence. The testing case is similar to that of the first set. We collect the user study results in Tab. V, from which we can see that randomly sampled target landmarks performs better than the two simplified methods. Compared to the single frame, our full method is considered as better in preserving the speaking style in 72% of all cases. Compared to the fixed sequence, our full method is considered as better in preserving the speaking style in 66% of all cases.

*4) User Study on Comparing to State-of-the-Art Methods:* Recent audio-driven dubbing methods also support audio-driven dubbing. We conduct a user study to compare our method with Audio2Obama, TBE, NVP, ATVGnet, Wav2Lip and EverybodyTalk. To compare with ATVGnet, Wav2Lip and EverybodyTalk, we apply the same training strategy, training dataset and training time to re-train them. We also collect 32 pairs of generated videos from our method and ATVGnet, Wav2Lip and EverybodyTalk. To compare with these methods, we collect their released results as our testing set. We ask each participant to evaluate which generated video is better in video quality, lip-sync accuracy and naturalness after watching the video pair. Similarly, we also compare our method with recent visual dubbing methods including Face2Face and DVP. The user study result is shown in Tab. VI. Though our method only requires very limited training data, the generated results are still competitive with the results produced by these recent methods designed for PGC production.

## V. Conclusion

We have proposed the first audio-driven dubbing framework that is designed specially to support ordinary video creators in User Generated Content (UGC) production. In particular, we have firstly explored an effective way to preserve speaking styles in audio-driven dubbing through the cross-modal AdaIN module and the Style Translation Network. We have also shown the possibility of driving an unseen face with its limited video data via the proposed semi-parametric Retrieval-based Video Renderer. Our method paves a way for ordinary content creators to edit talking videos in user generated content production. We hope our work will inspire more research in this field.

## Acknowledgment

The authors would like to thank the associate editor and the reviewers for their valuable comments and advice. Chen Change Loy would like to thank the cash and in-kind contribution from the industry partner(s).

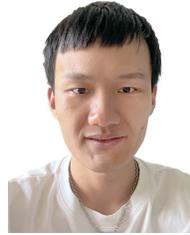

**Linsen Song** received the B.E. degree in electronic engineering from the University of Science and Technology Beijing in 2018. He is currently pursuing the Ph.D. degree with the National Laboratory of Pattern Recognition, Center for Research on Intelligent Perception and Computing, Institute of Automation, Chinese Academy of Sciences, and the School of Artificial Intelligence, University of Chinese Academy of Sciences, Beijing, China. His research interests include pattern recognition and computer vision.

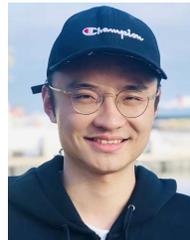

**Wayne Wu** received the Ph.D. degree from the BNRist Center, Department of Computer Science and Technology, Tsinghua University, Beijing, China. He is currently an Associate Director of Research and Development at SenseTime Group Inc. He is also an Adjunct Research Scientist at the Shanghai AI Laboratory, Shanghai, China. During his Ph.D. degree, he was a Visiting Ph.D. Student at the School of Computer Science and Engineering, Nanyang Technological University, Singapore. His research interests lie at the intersection of computer vision, computer graphics, and XR.

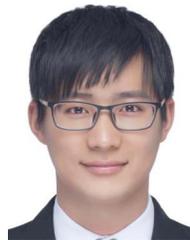

**Chaoyou Fu** received the B.E. degree in automation from Anhui University, Hefei, China, in 2017, and the Ph.D. degree in pattern recognition and intelligent systems from the National Laboratory of Pattern Recognition, Institute of Automation, Chinese Academy of Sciences, Beijing, China, in 2022. His research interests include pattern recognition and computer vision.

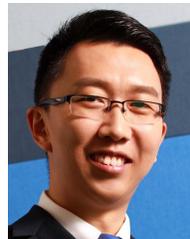

**Chen Change Loy** (Senior Member, IEEE) received the Ph.D. degree in computer science from the Queen Mary University of London in 2010. He is currently an Associate Professor with the School of Computer Science and Engineering, Nanyang Technological University, Singapore. He is also an Adjunct Associate Professor at The Chinese University of Hong Kong. Prior to joining NTU, he worked as a Research Assistant Professor at the MMLab, The Chinese University of Hong Kong, from 2013 to 2018. He was a Post-Doctoral Researcher at the Queen Mary University of London and Vision Semantics Ltd., from 2010 to 2013. His research interests include image/video restoration and enhancement, generative tasks, and representation learning. He serves as an Associate Editor of the IEEE TRANSACTIONS ON PATTERN ANALYSIS AND MACHINE INTELLIGENCE and *International Journal of Computer Vision*. He also serves/served as an Area Chair of major conferences, such as ICCV, CVPR, and ECCV.

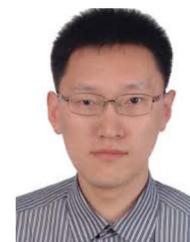

**Ran He** (Senior Member, IEEE) received the B.E. and M.S. degrees in computer science from the Dalian University of Technology in 2001 and 2004, respectively, and the Ph.D. degree in pattern recognition and intelligent systems from CASIA in 2009. Since September 2010, he has been joining NLPR, where he is currently a Full Professor. His research interests include information theoretic learning, pattern recognition, and computer vision. He serves as an Associate Editor of *Neurocomputing* (Elsevier) and serves on the program committee of several conferences. He is a fellow of the IAPR.